\documentclass[letterpaper]{article} 
\usepackage{aaai24}  
\usepackage{times}  
\usepackage{helvet}  
\usepackage{courier}  
\usepackage[hyphens]{url}  
\usepackage{graphicx} 
\usepackage{natbib}  
\usepackage{caption} 
\usepackage{algorithm}
\usepackage{algorithmic}
\usepackage{subfigure}
\usepackage{amsmath}
\usepackage{amssymb}
\usepackage{booktabs}
\usepackage{multirow}
\usepackage{dsfont}
\usepackage{color}
\usepackage{siunitx}
\usepackage{bm}
\usepackage{tabularx}
\usepackage{caption}
\usepackage{verbatim}
\usepackage{float}
\usepackage{pifont}
\usepackage{amsthm}
\usepackage{verbatim}
\usepackage{rotating} 
\usepackage{tabularx}
\usepackage{booktabs}
\usepackage{mathrsfs}
\usepackage[switch]{lineno}
\usepackage{newfloat}
\usepackage{listings}
\DeclareCaptionStyle{ruled}{labelfont=normalfont,labelsep=colon,strut=off} 
\floatstyle{ruled}
\newfloat{listing}{tb}{lst}{}
\floatname{listing}{Listing}
\title{SPGroup3D: Superpoint Grouping Network for Indoor 3D Object Detection}
\author{
    Yun Zhu\textsuperscript{\rm 1}, 
    Le Hui\textsuperscript{\rm 2}, 
    Yaqi Shen\textsuperscript{\rm 1}, 
    Jin Xie\textsuperscript{\rm 1}\thanks{Corresponding author.}
}
\affiliations{
    \textsuperscript{\rm 1} PCA Lab, 
    School of Computer Science and Engineering, Nanjing University of Science and Technology, China\\
    \textsuperscript{\rm 2} Shaanxi Key Laboratory of Information Acquisition and Processing, Northwestern Polytechnical University, China\\
    zhu.yun@njust.edu.cn,  huile@nwpu.edu.cn, syq@njust.edu.cn, csjxie@njust.edu.cn
}

\usepackage{bibentry}

\begin{document}

\maketitle

\begin{abstract}
Current 3D object detection methods for indoor scenes mainly follow the voting-and-grouping strategy to generate proposals. 
However, most methods utilize instance-agnostic groupings, such as ball query, leading to inconsistent semantic information and inaccurate regression of the proposals. 
To this end, we propose a novel superpoint grouping network for indoor anchor-free one-stage 3D object detection. 
Specifically, we first adopt an unsupervised manner to partition raw point clouds into superpoints, areas with semantic consistency and spatial similarity.
Then, we design a geometry-aware voting module that adapts to the centerness in anchor-free detection by constraining the spatial relationship between superpoints and object centers.
Next, we present a superpoint-based grouping module to explore the consistent representation within proposals. 
This module includes a superpoint attention layer to learn feature interaction between neighboring superpoints, and a superpoint-voxel fusion layer to propagate the superpoint-level information to the voxel level.
Finally, we employ effective multiple matching to capitalize on the dynamic receptive fields of proposals based on superpoints during the training.  
Experimental results demonstrate our method achieves state-of-the-art performance on ScanNet V2, SUN RGB-D, and S3DIS datasets in the indoor one-stage 3D object detection. Source code is available at https://github.com/zyrant/SPGroup3D.

\end{abstract}

\section{Introduction}
\begin{figure}[htbp]
	\includegraphics[width=1.00\linewidth]{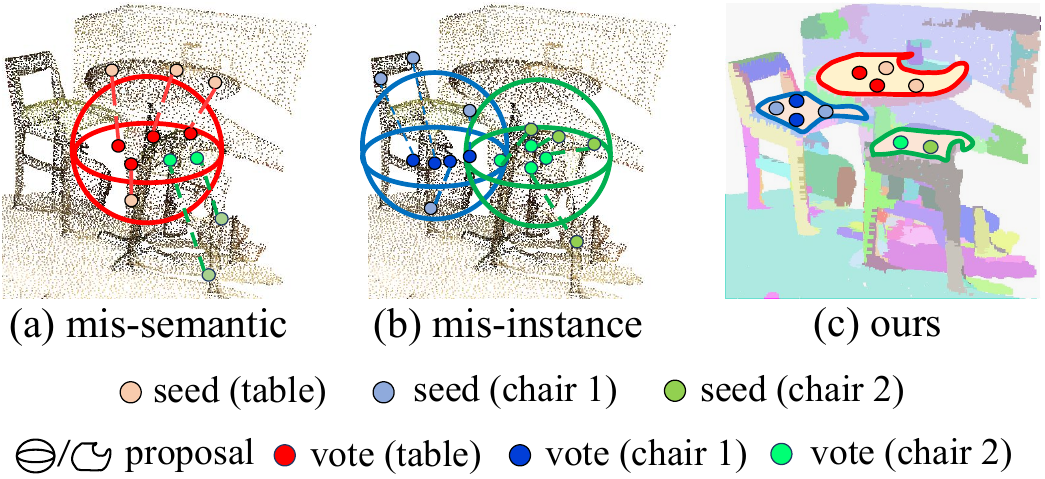}
	\caption{\text{Different types of grouping strategy}: (a) semantic-agnostic grouping; (b) semantic-aware grouping \cite{cagroup}; (c) our superpoint-based grouping. The proposal corresponding to the table is shown in red, and the two chairs are blue and green, respectively. For superpoint-based grouping, we represent different superpoints with different colors, and the features with the same superpoint will be grouped.}
	\label{fig:different types}
	\vskip -10pt
\end{figure}

As one of the basic tasks of 3D scene understanding, the goal of 3D object detection is to estimate the oriented 3D bounding boxes and semantic labels of objects in point clouds. It has been used in many application scenarios, such as autonomous driving, augmented reality, and robotics. Since indoor scenarios have more complex geometry and occlusion than outdoor scenarios \cite{sasa, svganet, te3d, seformer}, it remains challenging to cluster the scenes and get the proposals accurately.

Previous state-of-the-art 3D object detection methods mainly follow the bottom-up paradigm, involving two key components: voting and grouping. These components are used to learn the point-wise center offsets and aggregate the points that vote to instance-agnostic local region, respectively. 
The conventional voting tends to push all proposals close to the object centers. It may not suitable for FCOS-like \cite{fcos3d, fcaf3d} detection methods since the distances from the proposals to the object centers play a crucial role in determining the quality of proposals. 
Besides, the commonly used grouping module can be categorized into two types: semantic-agnostic and semantic-aware. The former, such as VoteNet and its subsequent works \cite{votenet, brnet, davotenet, cf3d}, groups the scenes based on the distance between coordinates. However, this approach fails in clustering indoor scenes when various instances are close but belong to different categories, as shown in Fig. \ref{fig:different types}(a). The latter one, such as CAGroup3D \cite{cagroup}, which additionally considers semantics, still suffers from a similar issue when instances of the same category are close to each other, as shown in Fig. \ref{fig:different types}(b). In summary, the conventional voting and the instance-agnostic grouping introduce more noise and outliers to the proposals, leading to performance bottlenecks in bottom-up methods.

In this paper, we propose a superpoint grouping network for indoor anchor-free one-stage 3D object detection (SPGroup3D) to solve these problems. Specifically, we first use a sparse convolution-based backbone to obtain the voxels, which include coordinates and features. 
Then, we introduce a geometry-aware voting module to adapt to the concept of centerness in the anchor-free detection methods \cite{fcos3d, fcaf3d}. For each voxel, by merging the seed parts and the vote parts, we can ensure the relative geometric relationships between proposals and object centers, making the model easier to filter out low-quality proposals.
Consequently, we address grouping problem by introducing superpoint \cite{spgraph}. As shown in Fig. \ref{fig:different types}(c), superpoint is a natural instance-aware local unit, consisting of adjacent points sharing similar semantics and spatiality. Based on superpoint, we construct a superpoint-based grouping module.
Superpoint-based grouping includes superpoint attention and superpoint-voxel fusion to facilitate superpoint-to-superpoint and superpoint-to-voxel information interactions, respectively. Superpoint attention is proposed to enhance the interaction between non-overlapping superpoints, adaptively exploring relevant superpoint features within their neighborhoods. The following superpoint-voxel fusion based on the sparse convolution is used for the interaction of voxels and superpoints. 
Finally, we employ multiple matching to assess the discrepancy between each superpoint-based proposal and the ground truth, and select positive samples during training. Extensive experiments indicate our method achieves state-of-the-art performance in the indoor one-stage 3D object detection task on three datasets, in terms of mAP@0.25 on ScanNet V2 (+1.1), SUN RGB-D (+1.2), and S3DIS (+2.5).

The main contributions are listed as follows:
\begin{itemize}
	\item We introduce a geometry-aware voting module, which preserves the relative geometry of superpoints in coordinate space, to adapt to anchor-free detection.
	\item We design a superpoint-based grouping module including superpoint attention and superpoint-voxel fusion to enable superpoint-to-superpoint and superpoint-to-voxel feature interactions, respectively. 
	\item We present a multiple matching strategy, which can discriminate between positive and negative samples of superpoint-based proposals during training.
\end{itemize}

\begin{figure*}[htbp]
	\centering
	\includegraphics[width=1.0\textwidth]{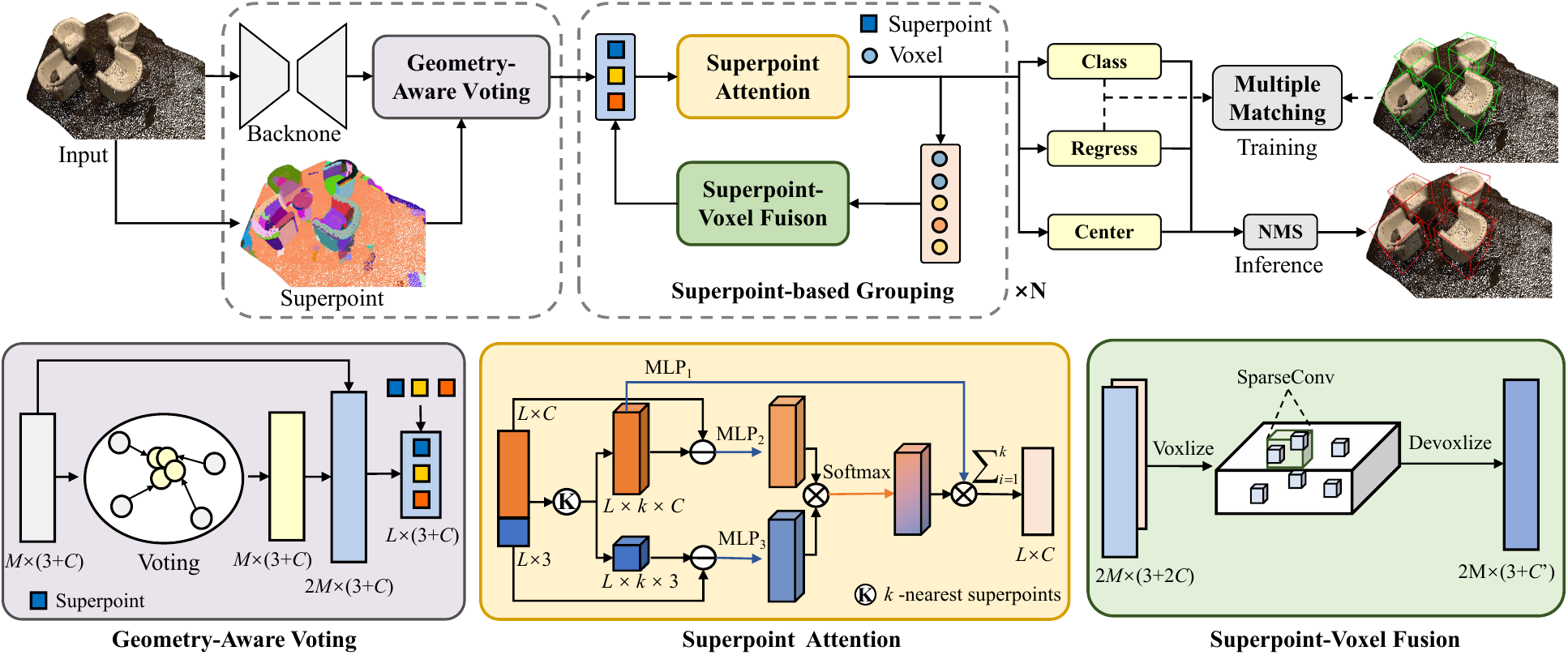}
	\caption{Framework of the superpoint grouping network for indoor 3D object detection (SPGroup3D). Given the input, we first extract seed voxels through the backbone. Subsequently, we construct geometry-aware voting to preserve the relative positions of the proposals and the object centers. Following this, based on superpoint-based grouping, we iteratively optimize the feature representations of the superpoints. Finally, multiple matching is employed to select positive samples during training and 3D NMS is applied to eliminate redundant proposals during the inference time.}
	\label{fig:SPGroup3D}
\end{figure*}

\section{Related Work}
\paragraph{Top-down 3D object detection.} Top-down models are primarily applied in outdoor autonomous driving scenarios. VoxelNet \cite{voxelnet} is a pioneering work that partitions the raw point clouds into voxels and realizes end-to-end 3D object detection in this field. PV-RCNN \cite{pvrcnn} argues that voxel quantization may lead to information loss and proposes a second stage for incorporating fine-grained features at the point level. For indoor 3D object detection, FCAF3D \cite{fcaf3d} introduces the first anchor-free model with fully sparse convolution and further proposes an improved version, TR3D \cite{tr3d}. 

\paragraph{Bottom-up 3D object detection.} Bottom-up methods, inspired by PointNet \cite{pointnet} and its variation \cite{pointnet++}, have gained widely attention for their ability to predict 3D bounding boxes from point clouds. VoteNet \cite{votenet} adopts deep hough voting to group features and generate proposals. Building upon this framework, MLCVNet \cite{mlcvnet} adopts different level context information to explore relationships between objects. H3DNet \cite{h3dnet} extends the key points prediction of centers in VoteNet, including the centers of bounding boxes, surfaces, and edges. VENet \cite{venet} designs a vote-weighting module to improve the voting process. BRNet \cite{brnet} proposes using clustered centers to predict the angles and spatial positions of entire objects. RBGNet \cite{rgbnet} introduces a ray-based feature grouping module to capture points on the object surface. CAGroup3D \cite{cagroup} leverages the powerful expressive capability of sparse convolution to enhance feature extraction and proposes a two-stage method. 
However, these methods essentially construct instance-agnostic grouping to generate proposals, which inevitably leads to the presence of multiple instance features within a proposal.

\paragraph{Superpoints.} Recently, superpoints have been introduced into other 3D point cloud tasks \cite{superlidar, spflow, spinter}. The concept of superpoints is initially proposed by SPG \cite{spgraph} and applied to 3D semantic segmentation. SSP \cite{ssp} proposes a learnable strategy for over-segmenting superpoints, while SPNet \cite{spoverseg} further extends the concept of superpoint oversegmentation to an end-to-end approach. In the domain of instance segmentation, SSTNet \cite{sstnet} constructs a superpoint tree network to aggregate superpoints with the same semantic information. GraphCut \cite{spcut} proposes a bilateral graph attention mechanism to generate precise instances using the superpoint graph cutting network. SPFormer \cite{spformer} utilizes learnable query to predict instance segmentation from superpoint features as a top-down pipeline. Although these methods have explored a variety of uses for superpoints, they have not directly enhanced the representation of superpoints nor made them more suitable for anchor-free detection. We employ a superpoint-based grouping to enhance the expression of superpoints and design a geometry-aware voting specifically for anchor-free detection to make the model easier to filter out low-quality proposals. 

\section{Method}

\subsection{Overview}
The overall architecture of our method is depicted in Fig. \ref{fig:SPGroup3D}. The input point cloud typically consists of $N$ points, which can be represented as $\{\bm{p}_i\}^{N}_{i=1}$, with $\bm{p}_i \in \mathbb{R}^6$. Each point has coordinates $x$, $y$, $z$ and colors $r$, $g$, $b$. Following previous method \cite{cagroup}, we first extract  $M$ high-resolution non-empty seed voxels from a sparse 3D backbone network \cite{minkowski}. Next, these voxels are passed through a geometry-aware voting module and several superpoint-based grouping modules to generate proposals. Finally, multiple matching is employed to choose positive proposals during training and 3D non-maximum suppression (NMS) is applied to remove redundant proposals during inference. 

\subsection{Geometry-Aware Voting} 
In anchor-free detection, we usually assign higher scores/ centernesses to the proposals/superpoints that are close to object centers and lower scores to proposals that are far away from object centers, and discard the proposals with low scores. 
However, traditional voting \cite{votenet} tends to push proposals close to the object centers, resulting in high scores, which makes it difficult to filter out low-quality proposals. In this paper, we present a geometry-aware voting to solve this problem. Geometry-aware voting preserves the relative position information from the proposals to the object centers in geometric space, enabling the model to allocate low scores for relatively distant proposals, thus making it easier to filter out low-quality proposals in post-processing.	

\begin{figure}[t]
	\vskip -10pt
	\centering
	\subfigure[]{
		\includegraphics[width=0.22\linewidth]{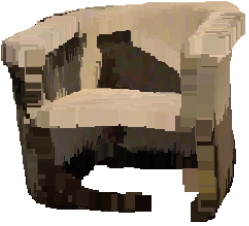}
		\label{fig:sofa_0}
	}
	\subfigure[]{
		\includegraphics[width=0.22\linewidth]{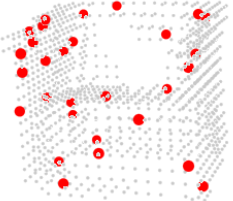}
		\label{fig:sofa}
	}
	\subfigure[]{
		\includegraphics[width=0.22\linewidth]{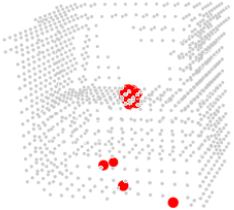}
		\label{fig:sofa_2}
	}
	\subfigure[]{
		\includegraphics[width=0.22\linewidth]{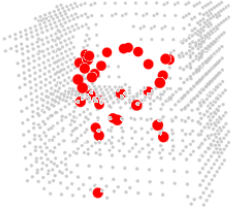}
		\label{fig:sofa_3}
	}
	\\ 
	\subfigure[]{
		\includegraphics[width=0.22\linewidth]{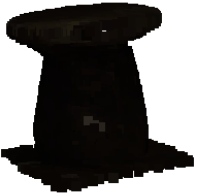}
		\label{fig:table_0}
	}
	\subfigure[]{
		\includegraphics[width=0.22\linewidth]{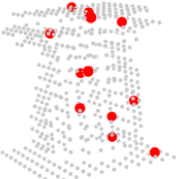}
		\label{fig:table}
	}
	\subfigure[]{
		\includegraphics[width=0.22\linewidth]{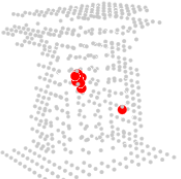}
		\label{fig:table_2}
	}
	\subfigure[]{
		\includegraphics[width=0.22\linewidth]{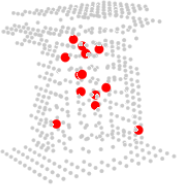}
		\label{fig:table_3}
	}
	\caption{The visualization of the locations of generated superpoints. The red points represent superpoints. To better represent the geometry of the object, we use gray points to indicate the voxels. (a) and (e) represent the input point clouds. (b) and (f) represent superpoints without voting. (c) and (g) represent superpoints with  traditional voting. (d) and (h) represent superpoints with geometry-aware voting.}
\end{figure}

Specifically, given $M$ high-resolution non-empty seed voxels $\{\bm{v}_i\}^{M}_{i=1}$ from the backbone,  where $\bm{v}_i = [\bm{v}_i^{c}; \bm{v}_i^{f}]$ with $\bm{v}_i^{c} \in \mathbb{R}^3$ and $\bm{v}_i^{f} \in \mathbb{R}^C$.The next key step is to gengerate proposals/superpoints. The most straightforward way to group voxels into superpoints is   to cluster $\{\bm{v}_i\}^{M}_{i=1}$ using offline pre-computed superpoints. However, as shown in Fig. \ref{fig:sofa} and (f), only using $\{\bm{v}_i\}^{M}_{i=1}$ will lead to the generated superpoints mainly existing on the surface of the objects, which are far away from the center of the instances, making it difficult to accurately regress \cite{votenet}.

Therefore, similar to VoteNet \cite{votenet}, we set a voting branch to learn the offsets of the voxels to the corresponding bounding box centers. Each seed voxel includes coordinate offset $\Delta \bm{v}_i^{c}\in\mathbb{R}^3$ and feature offset $\Delta \bm{v}_i^{f}\in\mathbb{R}^C$. The vote voxel $o_i$ is generated by adding the offsets as follows:
\begin{equation}
	\left\{\bm{o}_i \mid \bm{o}_i =[\bm{v}_i^{c}+\Delta \bm{v}_i^{c}; \bm{v}_i^{f}+\Delta \bm{v}_i^{f}] \right\}_{i=1}^M
\end{equation}
where the corresponding coordinate and feature of $\bm{o}_i$ are denoted by $\bm{o}_i^{c} \in \mathbb{R}^3$ and $\bm{o}_i^{f} \in \mathbb{R}^C$, respectively. The predicted offset $\Delta \bm{v}_i^{c}$ is explicitly supervised by a smooth-$\ell_1$ loss with the ground-truth displacement from the coordinate of seed voxel $\bm{v}_i$ to its corresponding bounding box center. Depicted in Fig. \ref{fig:sofa_2} and (g), the generated superpoints primarily emphasize the object centers with this traditional voting, weakening the geometry representation and relative position to object centers \cite{brnet}. 

This interference poses challenges to the post-processing in the anchor-free detection methods \cite{fcos3d, fcaf3d} since the superpoints belonging to the current object tend to gather around the object center. Hence, unlike VoteNet, which directly utilizes the vote voxels, we further merge the seed voxels. This merge processing can be formulated as follows:
\begin{equation}
	\left\{\bm{f}_i\mid \bm{f}_i = [(\bm{v}_i^{c}, \bm{o}_i^{c});  (\bm{v}_i^{f}, \bm{o}_i^{f})] \right\}_{i=1}^{2M}
\end{equation}
where $\bm{f}_i$ denotes the $i$-th voxel of $\{\bm{f}_i\}^{2M}_{i=1}$. This processing ensures that the generated superpoints remain within their corresponding objects while preserving their original relative position to the object centers, as illustrated in Fig. \ref{fig:sofa_3} and (h). 

\subsection{Sperpoint-based Grouping}
For the purpose to realize the consistent instance representation within the same proposals, we propose superpoint-based grouping to aggregate voxels into superpoints, consisting of superpoint attention and superpoint-voxel fusion.
In our setting, we iterate the superpoint-based grouping multiple times, \textit{i.e.}, 3. For each iteration, the output of superpoint attention is regarded as the corresponding output. Conversely, the output of superpoint-voxel fusion serves as the input for the next iteration of superpoint-based grouping.	

\paragraph{Superpoint Attention.} Superpoints are non-overlapping local regions of the input voxels, which become the bottleneck of the model perception, preventing dense prediction tasks like object detection from understanding the instances. In this paper, inspired by attention mechanism \cite{attention}, we propose a superpoint attention mechanism to enable feature interaction within neighboring local regions. The superpoint attention mechanism facilitates information propagation and integration, leading to a better understanding of the semantic and structural aspects of the instances. 

Specifically, given the $\{\bm{f}_i\}^{2M}_{i=1}$ obtained from the former module, we first group them into initial $L$ superpoints $\{\bm{s}_i\}^{L}_{i=1}$, where $\bm{s}_i = [\bm{s}_i^{c}; \bm{s}_i^{f}]$ with $\bm{s}_i^{c} \in \mathbb{R}^3$ and $\bm{s}_i^{f} \in \mathbb{R}^C$, using the superpoints. Subsequently, to accelerate the algorithm convergence and simplify its complexity, we apply the k-nearest neighbours ($k$-NN) algorithm to obtain the $k$ nearest neighbours based on coordinate space for each superpoint. The attention operation is then performed only within the $k$ nearest neighbours of each superpoint. These processes can be formulated as follows:
\begin{equation}
	\begin{aligned}
		\left\{{{\bm{s}_i}}\right\}^{L}_{i=1} &= Scatter\left\{{{\bm{f}_i}}\right\}^{2M}_{i=1}, \quad    
		\left\{{{\bm{n}_i}_j}\right\}^{k}_{j=1} = knn(\bm{s}_i)
	\end{aligned}
\end{equation}
where $Scatter$ and $\{{\bm{N}_i}_j\}^{k}_{j=1}$ represent Scatter\footnote{https://github.com/rusty1s/pytorch\_scatter} function and the $k$ nearest superpoints of $\bm{s}_i$,  respectively. Each superpoint in $\{{\bm{N}_i}_j\}^{k}_{j=1}$ can be represented by coordinate $\bm{n}_{ij}^{c} \in \mathbb{R}^3$ and feature $\bm{n}_{ij}^{f} \in \mathbb{R}^C$. 

Inspired by \cite{spcut}, we also explicitly calculate the similarity in both coordinate and feature space instead of embedding the coordinate into feature space. Different from their method that computes similarities between graphs for graph cutting in that realizes instance segmentation, our algorithm aims to facilitate interaction between superpoints to capture more contextual information. The specific steps are described as follows:
\begin{equation}
	\{\bm{w}_{ij}^{c}\}^{k}_{j=1} = \mathrm{MLP}(\bm{s}_i^{c} - 
	\{\bm{n}_{ij}^{c}\}^{k}_{j=1})
\end{equation}
\begin{equation}
	\{\bm{w}_{ij}^{f}\}^{k}_{j=1} = \mathrm{MLP}(\bm{s}_i^{f} - \{\bm{n}_{ij}^{f}\}^{k}_{j=1})	
\end{equation}
where $\{\bm{w}_{ij}^{c}\}^{k}_{j=1}$ and $\{\bm{w}_{ij}^{f}\}^{k}_{j=1}$ represent the wights in coordinate space and feature space of $k$ nearest superpoints, respectively. MLP indicates a fully-connected layer. It is noted that to guarantee the above process, we copy $\bm{s}_i^c$ and $\bm{s}_i^f$ $k$ times.

The fusion weights can be obtained by multiplying the coordinate space weights and feature space weights, followed by applying a softmax function to obtain normalized fusion weights.
Therefore, the fusion wight $w_{ij}$ for $i$-th superpoint and its $j$-th nearest superpoint can be formulated as:

\begin{equation}
	\bm{w}_{ij} = \frac{exp(\bm{w}_{ij}^{c} \times \bm{w}_{ij}^{f})} {\sum^{k}_{j=1} exp(\bm{w}_{ij}^{c} \times \bm{w}_{ij}^{f})}
\end{equation}

Finally, these fusion weights are multiplied by the superpoint features of the $k$-nearest neighbours to obtain updated superpoint feature, while the coordinate of the superpoint remains unchanged. The step can be formulated as:
\begin{equation}
	\bm{a}_i^f = \sum^{k}_{j=1}({\bm{w}_i}_j \times \mathrm{MLP}(\bm{n}_{ij}^{f})), \quad
	\bm{a}_i^c = \bm{s}_i^c
\end{equation}
where $\bm{a}_i^f$ and $\bm{a}_i^c$ denote one of the features and coordinates of the updated superpoints in $\{\bm{a}_i\}^{L}_{i=1}$, respectively. In addition to above steps, similar to the original attention algorithm, we also incorporate residual and normalization operations at last. 

\paragraph{Superpoint-Voxel Fusion.} From voxels to superpoints is a quantification process, inevitably leading to a loss of fine-grained features. Therefore, we propose superpoint-voxel fusion based on sparse convolution to achieve the interaction of coarse-grained superpoints and fine-grained voxels. 

First, we need to broadcast the superpoint features to match the size of the merge voxels $\{\bm{f}_i\}^{2M}_{i=1}$ using the superpoints. The superpoint features after the broadcast are denoted as $\{\bm{h}_i^f\}^{2M}_{i=1}$. Then, we concatenate the features from $\{\bm{f}_i\}^{2M}_{i=1}$  and $\{\bm{h}_i\}^{2M}_{i=1}$ to obtain the initial fusion voxel features $\{\bm{g}_i^{f}\}^{2M}_{i=1}$. The step can be formulated as:

\begin{equation}
	\{{\bm{h}_i^f}\}^{2M}_{i=1} = \operatorname{Broadcast}\left(\{{\bm{a}_i^f}\}^{L}_{i=1}\right)
\end{equation}
\begin{equation}
	\{{\bm{g}_i^{f}}\}^{2M}_{i=1} = \operatorname{Concat}\left(\{{\bm{f}_i^{f}}\}^{2M}_{i=1}; \{{\bm{h}_i^{f}}\}^{2M}_{i=1}\right)
\end{equation}
where $\bm{g}_i^{f} \in \mathbb{R}^{2C}$ is one the feature of fusion voxels $\{\bm{g}_i\}^{2M}_{i=1}$. Besides, the coordinates of the fusion voxels are the same as those of the merge voxels. Next, based on the coordinates, we will voxelize fusion voxels again with a voxel size equal to the resolution of output voxels from the backbone. The revoxlized fusion voxels pass through an SPFFN, consisting of a sparse convolution layer, a normalization layer, and an activation layer to obtain the final output. This process can be formulated as follows:
\begin{equation}
	\{{\bm{g}_i^{'}}\}^{M}_{i=1} = \mathrm{SPFFN}\left(\operatorname{Voxelize}(\{{\bm{g}_i}\}^{2M}_{i=1})\right)
\end{equation}
where $\bm{g}_i^{'}$ denotes one of updated fusion voxels. Finally, $\{{\bm{g}_i^{'}\}}^{M}_{i=1}$ will be converted back to input-level by the sparse tensor matrix mapping and then get the refined voxels $\{\bm{f}_{i+1}\}^{2M}_{i=1}$ for next iteration. 

\subsection{Multiple Matching and Loss Function}
Previous anchor-free indoor 3D detection methods \cite{fcaf3d, cagroup} rely on constructing proposals using regular neighbourhoods, \textit{i.e.}, voxels. Assume we consider the receptive field as a criterion for determining the importance of the current proposal, proposals within the same features level would be treated equally. Based on this pattern, they could directly set positive and negative samples by judging the distance between the location of the proposals and the center of the bounding boxes. 

However, this approach is not suitable for proposals generated by superpoints due to the fact that the proposals in our method have dynamic receptive fields. We must improve the matching strategy to select multiple positive samples for each ground truth. Here, to ensure consistency between training and registration, inspired by DETR \cite{detr}, we directly adopt the loss during the training process as the cost function for matching and consider the cost of both classification and regression simultaneously. It is worth mentioning that our approach does not require Hungarian Matching and assigns multiple samples for each ground truth,  which is different from Bipartite Matching in DETR.
$Cost_{ik}$ is to evaluate the similarity of the $i$-th proposal and the $k$-th ground truth. As defined by the following formula:
\begin{equation}
	Cost_{ik} = -\lambda_{cls}cost_{ik}^{cls} - \lambda_{reg}cost_{ik}^{reg}
\end{equation}
where $cost_{ik}^{cls}$ and $cost_{ik}^{reg}$ represent the focal cost function \cite{detr} and DIoU cost function \cite{DIOU}, respectively. $\lambda_{cls}$ and $\lambda_{reg}$ are the corresponding coefficients for each term. Finally, we directly select the top-$r$ (\textit{i.e.}, $r$=18) proposals with the minimum cost as positive samples for each ground truth, while the rest are considered as negative. In our experiments, only the proposals within the bounding box are considered, and both $\lambda_{cls}$, $\lambda_{reg}$ are set to 1.

After assignment, our method is trained from scratch with voting loss $\mathcal{L}_{{vote}}$, centerness loss $\mathcal{L}_{cntr}$, bounding box estimation loss $\mathcal{L}_{box}$, and classification loss $\mathcal{L}_{cls}$ for object detection, which are formulated as follows:
\begin{equation}
	L = \beta_{vote}\mathcal{L}_{{vote}} + \beta_{cntr}\mathcal{L}_{cntr} \\ 
	+\beta_{box}\mathcal{L}_{box} + \beta_{cls}\mathcal{L}_{cls}
\end{equation}

$\mathcal{L}_{{vote}}$ is a smooth-$\ell_1$ loss for predicting the center offset of each voxel. In terms of proposal generation, $\mathcal{L}_{cntr}$, $\mathcal{L}_{box}$, and $\mathcal{L}_{cls}$ utilize Cross-Entropy loss, DIOU loss \cite{DIOU}, and Focal loss \cite{focal} to optimize object centerness, bounding box prediction, and classification, respectively. $\beta_{vote}$, $\beta_{cntr}$, $\beta_{box}$, and $\beta_{cls}$ represent the corresponding coefficients and are set to 1.

\begin{table*}[!htb]
	\centering
	\resizebox{\textwidth}{!}{
		\begin{tabular}{c|c|cc|cc|cc}
			\toprule
			\multirow{2}{*}{Methods} & \multirow{2}{*}{Presented at}
			& \multicolumn{2}{c|}{ScanNet V2} & \multicolumn{2}{c|}{SUN RGB-D} & \multicolumn{2}{c}{S3DIS} \\
			& & $\text{mAP}@0.25$ & $\text{mAP}@0.5$ & $\text{mAP}@0.25$ & $\text{mAP}@0.5$ & $\text{mAP}@0.25$ & $\text{mAP}@0.5$ \\
			\midrule
			\midrule
			VoteNet \cite{votenet} & ICCV’19 & 58.6 & 33.5 & 57.7 & - & - & - \\ 
			3D-MPA \cite{3d_mpa} &CVPR’20& 64.2 & 49.2  & - & - & - & - \\
			HGNet \cite{hgnet}  &CVPR’20& 61.3 & 34.4  & 61.6 & - & - & - \\
			MLCVNet \cite{mlcvnet}  &CVPR’20& 64.5 & 41.4  &  59.8 & - & - & -\\
			GSDN (Gwak et al. 2020) &ECCV’20& 62.8 & 34.8 & - & - & 47.8 & 25.1 \\
			H3DNet \cite{h3dnet}  &ECCV’20& 67.2 & 48.1  & 60.1 & 39.0 & - & - \\
			BRNet \cite{brnet} &CVPR’21& 66.1 & 50.9   & 61.1 & 43.7 & - & -\\
			3DETR (Misra et al. 2021) &ICCV’21& 65.0 & 47.0  & 59.1 & 32.7 & - & -\\
			VENet \cite{venet} &ICCV’21& 67.7 & -   & 62.5 & 39.2 & - & -\\
			GroupFree \cite{groupfree} &ICCV’21& 69.1 (68.6) & 52.8 (51.8)  & 63.0 (62.6) & 45.2 (44.4) & - & -\\
			RBGNet \cite{rgbnet} &CVPR’22& 70.6 (69.6) & 55.2 (54.7)  & 64.1 (63.6) & 47.2 (46.3) & - & -\\
			HyperDet3D \cite{hyperdet3d} &CVPR’22& 70.9 & 57.2  & 63.5 & 47.3 & - & -\\
			FCAF3D (Rukhovich et al. 2022) &ECCV’22& 71.5 (70.7) & 57.3 (56.0)  & 64.2 (63.8) & \textbf{48.9} (48.2) & 66.7 (64.9) & 45.9 (43.8)\\
			CAGroup3D* \cite{cagroup} &NeurIPS’22& 73.2 & 57.1  & - & - & - & -\\ 
			\midrule
			SPGroup3D (ours) &-& \textbf{74.3} (73.5) & \textbf{59.6} (58.3) & \textbf{65.4} (64.8) & 47.1 (46.4) & \textbf{69.2} (67.7) & \textbf{47.2} (43.6) \\
			\bottomrule
	\end{tabular}}
	\caption{3D detection results on validation of ScanNet V2, SUN RGB-D, and S3DIS datasets. The main comparison is based on the best results of multiple experiments between different methods, and the average value of 25 trials is given in brackets. Since we only focus on one-stage methods in this paper, we report the one-stage result from the ablation study of the origin paper \cite{cagroup} for fair comparison, dubbed as `` CAGroup3D* " .}
	\label{tab:main result}
\end{table*}

\begin{table}[t]
	\centering
	\resizebox{\linewidth}{!}{
		\begin{tabular}{ccc|cc}
			\toprule
			Geometry & Group &  Match &  $\text{mAP}@0.25$ & $\text{mAP}@0.5$ \\
			\midrule
			\midrule
			& & & 68.2 & 53.2  \\
			\checkmark & & & 71.3 & 57.0  \\
			\checkmark & \checkmark & & 72.7 & 57.9  \\
			\checkmark & \checkmark & \checkmark & \textbf{73.5} & \textbf{58.3}\\
			\bottomrule
	\end{tabular}}
	\caption{Ablation study of key components including geometry-aware voting (Geometry), superpoint-based grouping (Group), and multiple matching (Match).}
	\label{tab:main component}

\end{table}

\section{Experiments}
\subsection{Datasets and Evaluation Metric}
Our SPGroup3D is evaluated on three indoor challenging 3D scene datasets, \textit{i.e.}, ScanNet V2 \cite{scannet}, SUN RGB-D \cite{sunrgbd}, S3DIS \cite{s3dis}. For all datasets, we follow the standard data splits adopted in \cite{votenet} and \cite{gsdn}.

\textbf{ScanNet V2.} ScanNet is a richly annotated dataset that provides a comprehensive collection of 3D indoor scans, with annotated 3D rebuilt indoor scenes and bounding boxes for the 18 object categories. The dataset is divided into 1,201 training samples, with the remaining 312 used for validation.

\textbf{SUN RGB-D.} SUN RGB-D is a widely recognized dataset designed for 3D object detection in indoor environments. This dataset is divided into approximately 5,000 training and 5,000 validation samples. Each sample is annotated with oriented 3D bounding boxes and per-point semantic labels, covering 37 different object categories. Following the approach in \cite{votenet}, we select 10 categories.

\textbf{S3DIS.} S3DIS is a comprehensive 3D indoor dataset that comprises 3D scans from 272 rooms across 6 buildings, with annotations for both 3D instances and semantic categories. We follow the standard division, where Area 5 is used for validation, while the remains are the training subset.

We evaluate all the experiment results by a standard evaluation protocol \cite{votenet,fcaf3d}, which uses mean average precision (mAP) with different IoU thresholds, \textit{i.e.}, 0.25 and 0.5.

\subsection{Implementation Details} 
We set the voxel size as 0.02$m$ for all datasets. For the backbone, we use the same backbone introduced in \cite{cagroup} as the voxel feature extractor and the voxel size of output high-resolution is 0.04$m$. In terms of the superpoint-based grouping, we set the iteration number to 3 and the neighbour number to 8. Moreover, following the setting in FCAF3D \cite{fcaf3d}, we set the number of positive samples to 18 in multiple matching. 

\begin{figure}[t]
	\subfigure[iteration number]{
		\includegraphics[width=0.47\linewidth]{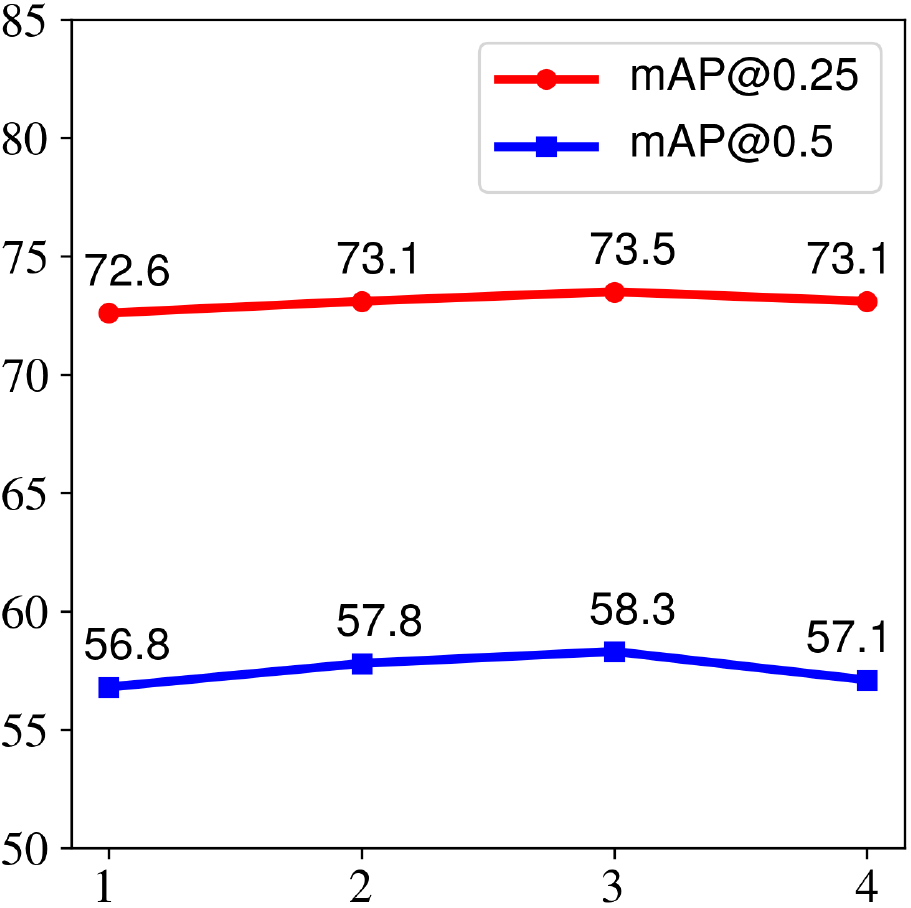}
		\label{fig:iter}
	}
	\subfigure[neighbour number]{
		\includegraphics[width=0.47\linewidth]{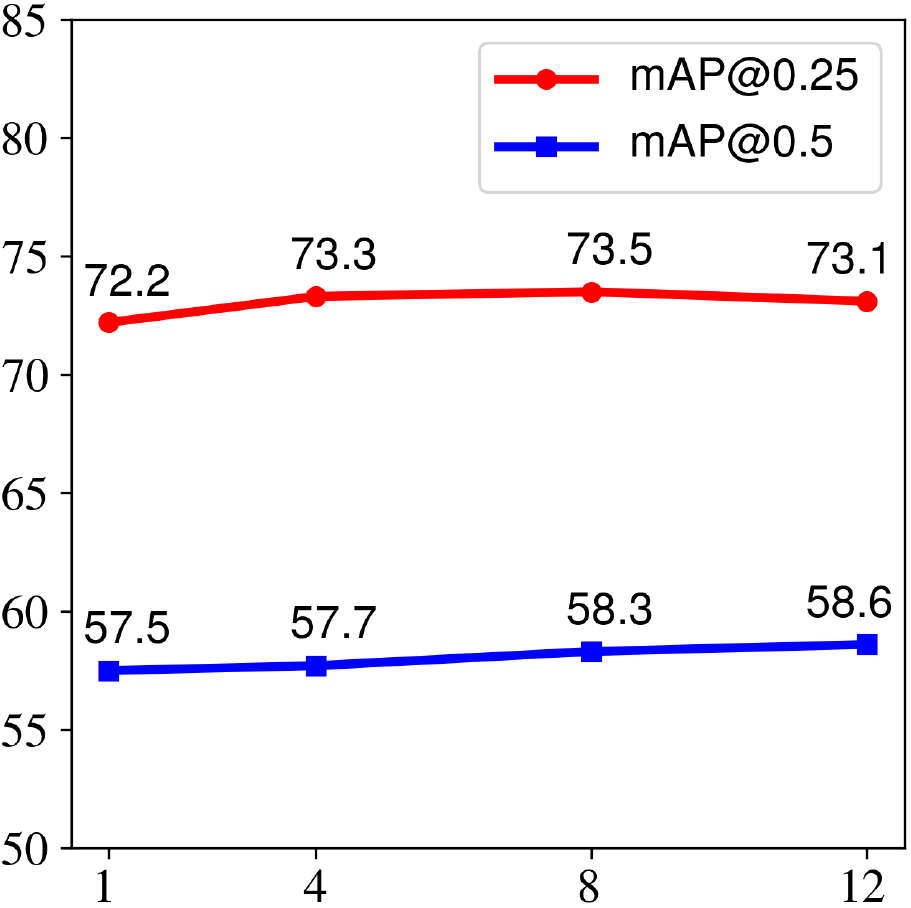}
		\label{fig:k}
	}
	\caption{Comparison of performance results for different iteration numbers and neighbour numbers.}
\end{figure}

In our experiments, we train the model in an end-to-end manner using the MMdetection3D framework \cite{mmdet3d}. Following the approach in \cite{fcaf3d, tr3d}, we employ the AdamW optimizer \cite{adam} with batch size, initial learning rate, and weight decay set to 4, 0.001, and 0.0001, respectively. Training is performed for 15 epochs on each dataset, with a learning rate decay by a factor of 10 at the 9-th and 12-th epochs. The experiments are conducted on four NVIDIA RTX 3090 GPUs. We follow the evaluation scheme from \cite{groupfree}, which involved training the models five times and testing each trained model five times. The reported results include the best and average performance across all results.

\begin{figure*}[htbp]
	\centering
	\includegraphics[width=1.0\textwidth]{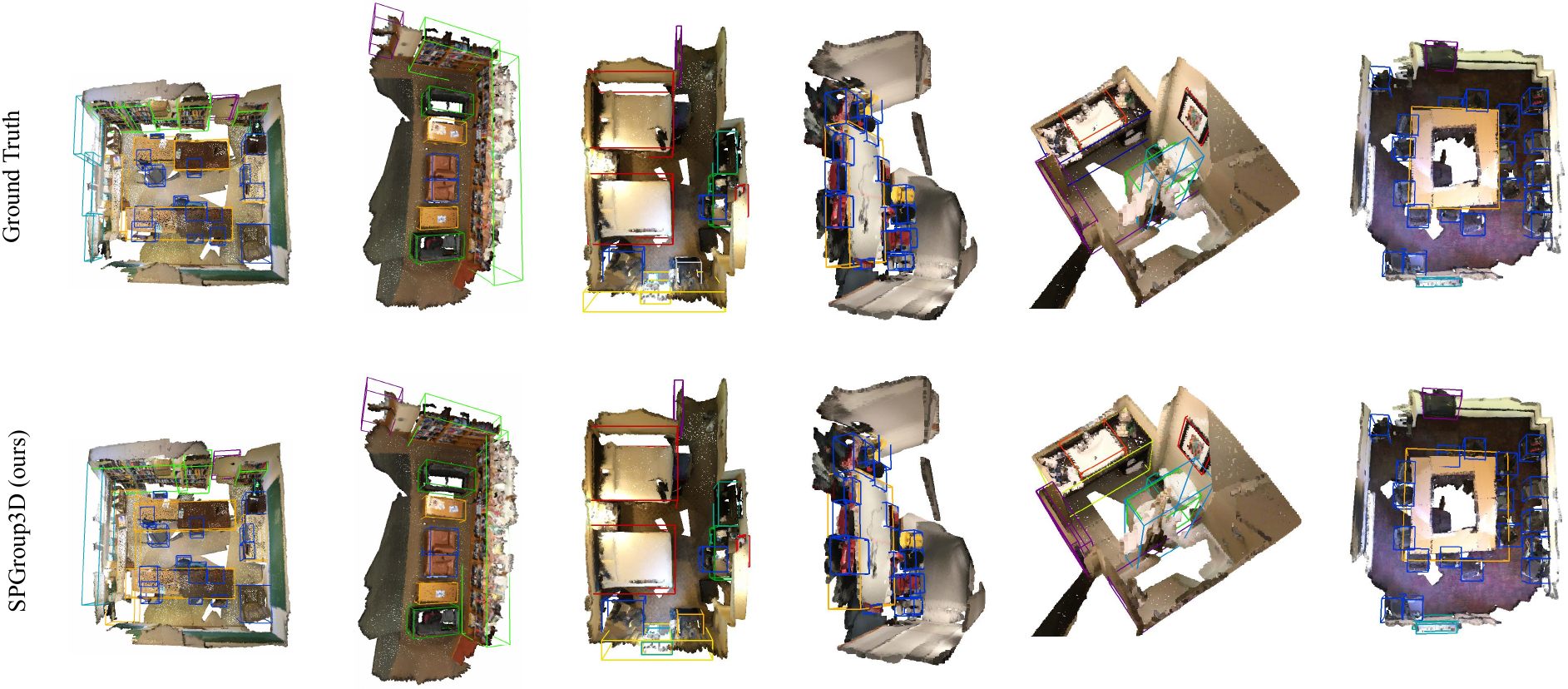}
	\caption{Qualitative results on validation of ScanNet V2. Different classes are indicated by bounding boxes in different colors.}
	\vskip +10pt
	\label{fig:scannet_show}
	
\end{figure*}

\subsection{Benchmarking Results}
We compare our method with the recent state-of-the-art 3D detection methods on ScanNet V2 \cite{scannet}, SUN RGB-D \cite{sunrgbd} and S3DIS \cite{s3dis} benchmarks. As indicated in Tab. \ref{tab:main result}, SPGroup3D outperforms the previous state-of-the-art methods in almost all metrics. In terms of mAP@0.25, our method achieves 1.1, 1.2, and 2.5 improvements over the previous state-of-the-art methods on ScanNet, SUN RGB-D and S3DIS, respectively. Regarding mAP@0.5, our method shows 2.3 and 1.3 improvements on ScanNet and S3DIS. The visualization of 3D Object detection with predicted bounding boxes on Scannet V2 is shown in Fig. \ref{fig:scannet_show}.

\subsection{Ablation Study}
We conduct extensive ablation studies on the validation set of ScanNet V2 to analyze individual components of our proposed method.

\paragraph{Effect of different components of SPGroup3D.}We first ablate the effects of different components of SPGroup3D. As seen in Tab. \ref{tab:main component}, the base model ($1{st}$ row) is the fully sparse convolutional VoteNet-style \cite{votenet, cagroup} model. Comparing the $1{st}$ and $2{nd}$ row, we introduce a geometry-aware voting (dubbed as ``Geometry"), eliminate superpoint-based grouping and multiple matching, and directly group voxels into superpoints. The results of this variant model are significantly improved from 68.2 to 71.3, and mAP@0.5 from 53.2 to 57.0. Comparing the $2{nd}$ and $3{rd}$ row, by adding several superpoint based groupings (dubbed as ``Group"), the performance of this variant further improved, \textit{i.e.}, mAP@0.25 and mAP@0.5, 71.3 to 72.7 and 57.0 to 57.9, respectively. In conjunction with multiple matching (dubbed as ``Match"), from the $4{th}$ row, the result of mAP@0.25 increased from 72.7 to 73.5, and mAP@0.5 from 57.9 to 58.3. These experiments demonstrate the effectiveness of our geometry-aware voting, superpoint-based grouping and multiple matching.

\begin{table}[!t]
	\centering
	\begin{tabular}{c|cc}
		\toprule
		Settings  &$\text{mAP}@0.25$ & $\text{mAP}@0.5$ \\
		\midrule
		\midrule
		Semantic-agnostic & 68.2 &53.2    \\
		Semantic-aware & 73.2 & 57.1  \\
		SPGroup3D (ours) & \textbf{73.5} & \textbf{58.3}  \\
		\bottomrule
	\end{tabular}
	\caption{Comparison with other grouping-based strategies.}
	\label{tab:grouping}
\end{table}

\paragraph{Effect of superpoint based grouping.}In the superpoint-based grouping, there are two hyperparameters: iteration number ($iter$) and neighbour number ($k$). According to Fig. 4, we can see that the model is insensitive to changes in hyperparameters. Fig. \ref{fig:iter} shows that the best results are obtained at an iteration number of 3 (73.5 on mAP@0.25 and 58.3 on mAP@0.5). Therefore, in our experiments, we set the number of iterations to 3. We choose several values for the number of neighbours: 1, 4, 8, and 12, as shown in Fig. \ref{fig:k}. The results of mAP@0.25 and mAP@0.5 are 72.2, 57.5 (k=1), 73.3, 57.7 (k=4), 73.5, 58.3 (k=8) and 73.1, 58.6 (k=12), respectively. In order to strike a balance between performance and efficiency, this paper chooses 8 as the number of neighbours. 

To further demonstrate the effect of superpoint-based grouping, we compare it with other grouping strategies. In Tab. \ref{tab:grouping}, ``semantic-agnostic" and ``semantic-aware" correspond to VoteNet-style model and the one-stage CAGroup3D \cite{cagroup}, respectively. Our method achieve more reliable detection results (73.5 for mAP@0.25, 58.3 for mAP@0.5) compared to these methods. This proves that instance-aware proposals produced by superpoint-based grouping lead to better detection.

\paragraph{Effect of geometry-aware voting.} In our geometry-aware voting, we need to preserve both features before and after voting. Here we study its impact.
Only using the features before voting achieves 72.0 for mAP@0.25 and 55.1 for mAP@0.5. The variant model only with the features after voting obtains 70.9 for mAP@0.25 and 56.2 for mAP@0.5, and with both features can achieve 73.5 for mAP@0.25 and 58.3 for mAP@0.5. It can be observed that by utilizing both features, our model achieves improvement. This shows that preserving the geometric distribution of superpoints in coordinate space is beneficial to the anchor-free method. Additionally, we claim that our model is robust to incorrectly partitioned superpoints, which usually appear at the edges of objects and are more likely to generate low-quality proposals. With geometry-aware voting, the position of superpoints relative to the center of objects remains unchanged, allowing us to easily use centerness scores to filter out low-quality proposals during post-processing.

\section{Conclusion}
In this paper, we proposed a novel end-to-end one-stage method, SPGroup3D, for indoor 3D object detection. SPGroup3D first utilizes geometry-aware voting to refine the positions of superpoints and then employs superpoint-based grouping to group the bottom-up latent features of voxels into superpoints. During the training phase, multiple matching is used to select positive superpoint-based proposals. Extensive experiments on ScanNet V2, SUN RGB-D, and S3DIS benchmarks demonstrate that the proposed method achieves state-of-the-art performance on the indoor one-stage 3D object detection task.

\section{Acknowledgments}
This work was supported by the National Science Fund of China (Grant Nos. 62276144, 62306238) and the Fundamental Research Funds for the Central Universities.

\bibliography{aaai24}

\clearpage
\appendix

\section{Overview}

This supplementary material provides more details on architecture, ablation study, per-category evaluation, and visualization of our method. Specifically, in Sec. \ref{1}, we provide network architecture details about the backbone, voting branch, superpoint-based grouping, and detection head.  In Sec. \ref{2}, we discuss more ablation studies. In Sec. \ref{3}, we provide per-category evaluation results and more visualization of our proposed method.

\section{Network Architecture} \label{1}
The backbone network is BiResNet \cite{cagroup}, which consists of two branches, one for ResNet18 \cite{gsdn}, with downsampling rates of 2, 4, 8, and 16, and the other is used to store high-resolution feature maps with a resolution of 1/4 of the input 3D voxels. All convolutional layers are followed by batch normalisation and ReLU activation function. The final output is upsampled to 1/2 of the input voxel. The output of the backbone network is a 64-dimensional voxel latent feature.

For the voting branch, we set three convolution layers. All convolutional layers are followed by batch normalisation and ELU activation function, except the last convolutional layer. The output of the voting branch is a 67-dimensional voxel latent feature, where the 3-dimensional feature is used to compute the coordinate offset and the rest 64-dimensional feature for feature offset.

For the superpoint-based grouping, the initial merge voxel features are added to the normalized coordinates corresponding to the superpoint coordinates and the superpoint coordinates to get 70-dimensional hidden features. For three rounds of iteration, the output dimensions of superpoint-based grouping are 64, 128, and 128, respectively. The final output will concatenate these features and the 70-dimensional features to obtain 390-dimensional features.

For the head, two shared fully-connected layers are first set up to enable information exchange between different channels. Each layer is followed by a layer normalisation and a ELU activation function. The output of the last activation function is sent to the classification head, regression head and centerness head to get the final prediction.

\section{Ablation Study} \label{2}
We conduct more studies on the validation set of ScanNet V2 to analyze different components of our proposed method. 

\begin{figure}[t]
	\centering
	\includegraphics[width=1\linewidth]{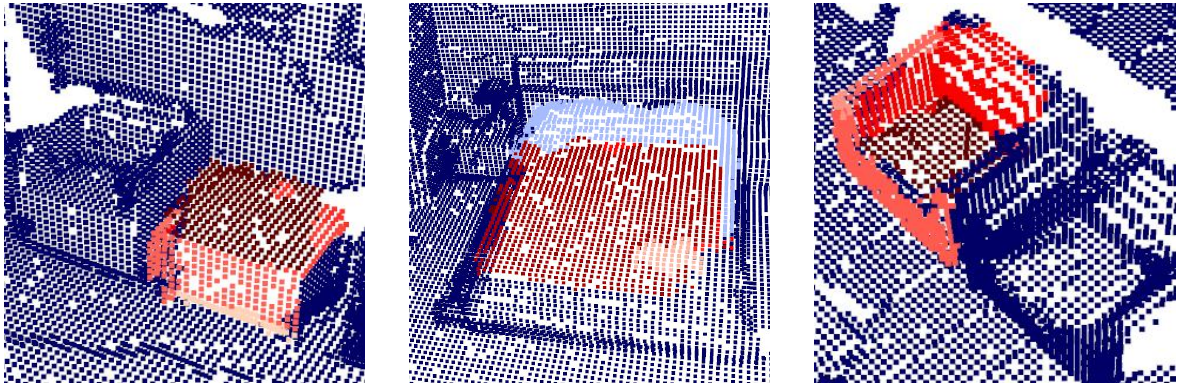}
	\caption{Attention maps in superpoint attention. For clear visualization, we return the superpoints to the voxel level and remove the vote voxels. The color indicates attention values: red for high and dark blue for low.}
	\label{fig:attn}
	\vskip -10pt
\end{figure}

\paragraph{Effect of superpoint attention.} In order to study the effect of superpoint attention, we visualize the attention maps. As shown in Fig. \ref{fig:attn}, attention weights almost cover the whole instance, which proves that our design is effective.

\paragraph{Effect of sparse conv.} Compared with the fully connected layer, sparse convolution can provide more features interaction. Tab. \ref{tab:sparse} shows the results of using a fully-connected layer and sparse convolution in superpoint-voxel fusion. Since sparse convolution achieves better results, we chose it as the default set in our experiment. 

\paragraph{More possible combinations of the key components.} There are three key components in this paper including geometry-aware voting (dubbed as ``Geometry"), superpoint-based grouping (dubbed as ``Group"), and multiple matching (dubbed as ``Match"). In Tab. \ref{tab:diff combi}, We list more results of different combinations of the key components, including the combination of Geometry and Match, and the combination of Group and Match. It can be seen that our well-designed components can still achieve performance improvement, which demonstrates the robustness and effectiveness of our method.

\begin{table}[h]
	\centering{
		\begin{tabular}{c|cc}
			\toprule
			& $\text{mAP}@0.25$ & $\text{mAP}@0.5$ \\
			\midrule
			\midrule
			MLP & 72.2 & 57.4  \\
			Sparse Conv & \textbf{73.5} & \textbf{58.3} \\
			\bottomrule
	\end{tabular}}
	\caption{Effect of sparse convolution.}
	\label{tab:sparse}
	\vskip -10pt
\end{table}

\begin{table}[h]
	\centering
	\resizebox{\linewidth}{!}{
		\begin{tabular}{ccc|cc}
			\toprule
			Geometry & Group &  Match &  $\text{mAP}@0.25$ & $\text{mAP}@0.5$ \\
			\midrule
			\midrule
			& & & 68.2 & 53.2  \\
			\checkmark & & \checkmark & 71.8 & 57.6  \\
			& \checkmark & \checkmark & 70.9 & 56.2  \\
			\checkmark & \checkmark & \checkmark & \textbf{73.5} & \textbf{58.3}\\
			\bottomrule
	\end{tabular}}
	\caption{Ablation study of different combinations of key components.}
	\label{tab:diff combi}
	\vskip -10pt
\end{table}

\section{More Results} \label{3}

\begin{figure*}[h]
	\centering
	\includegraphics[width=1.0\textwidth]{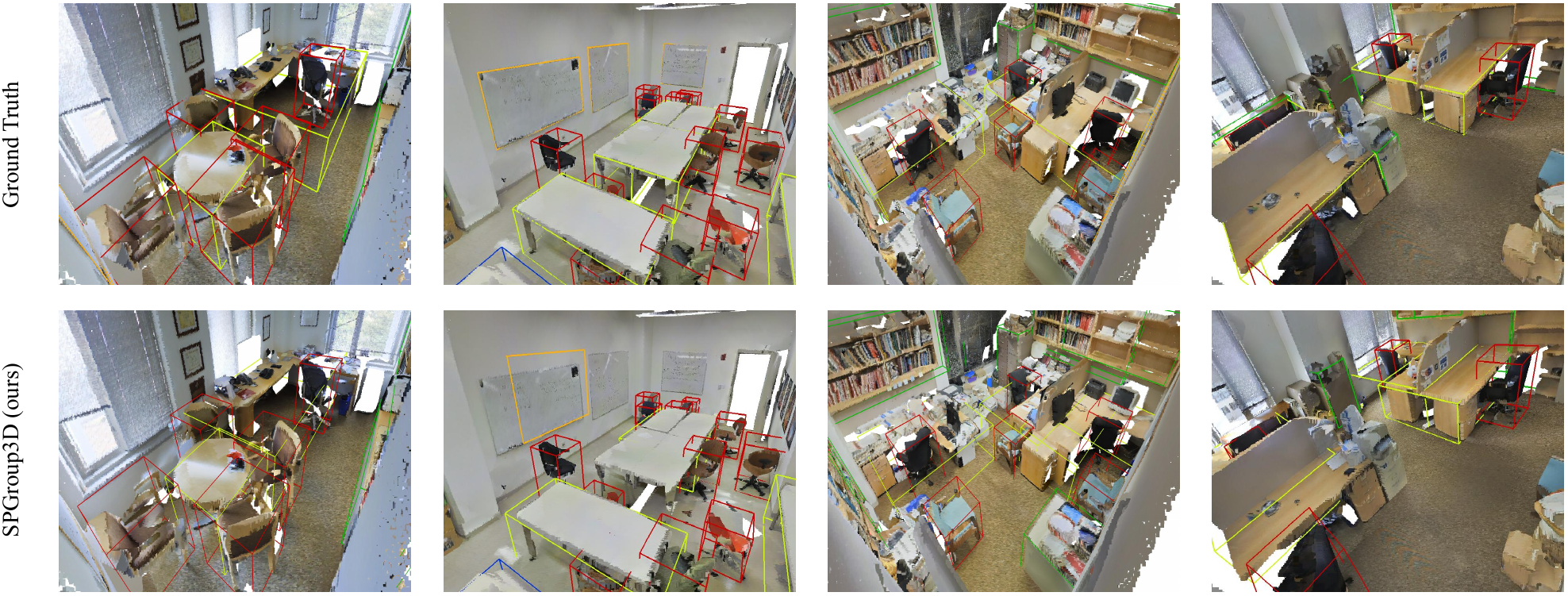}
	\caption{Qualitative results on S3DIS. Different classes are indicated by bounding boxes in different colors.}
	\label{fig:s3dis_show}
	\vskip -5pt
\end{figure*}

\begin{figure*}[h]
	\centering
	\includegraphics[width=1.0\textwidth]{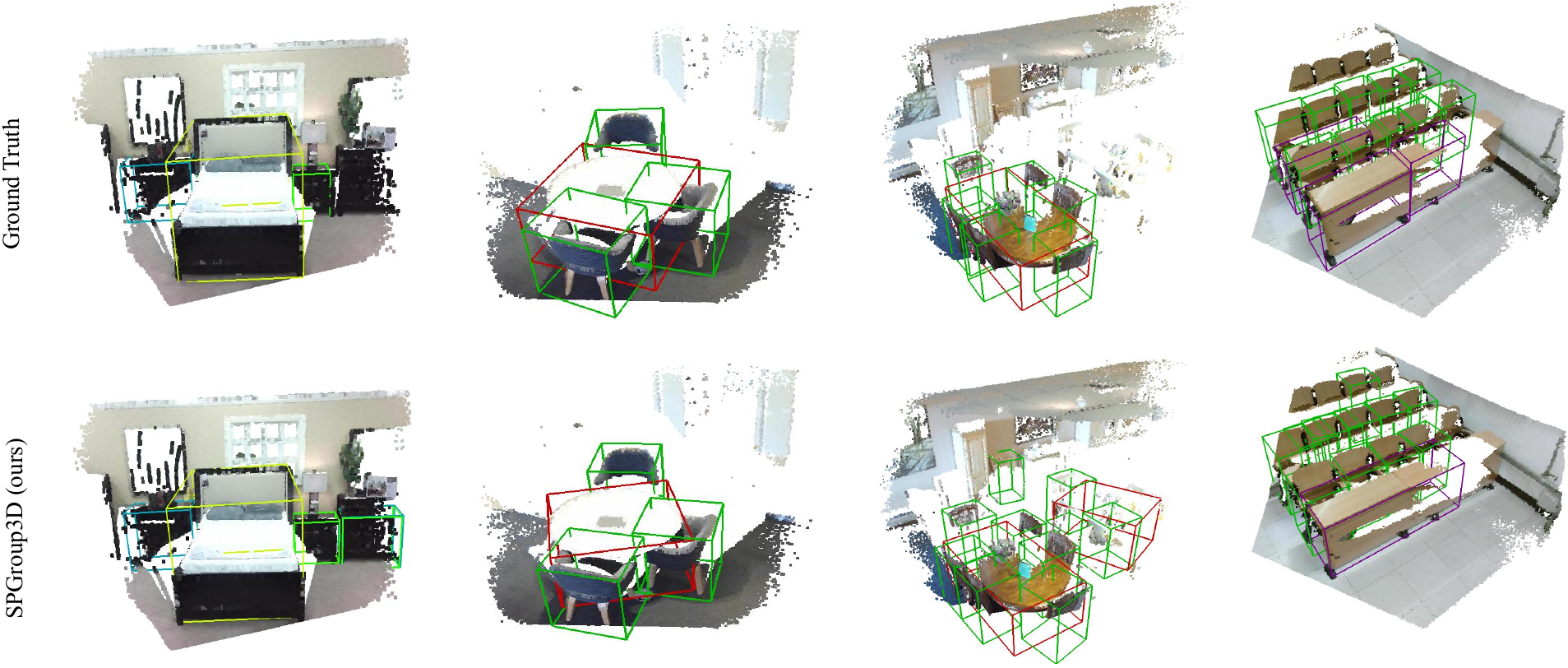}
	\caption{Qualitative results on SUN RGB-D. Different classes are indicated by bounding boxes in different colors.}
	\label{fig:sunrgbd_show}
	\vskip -10pt
\end{figure*}

\paragraph{More quantitative Results.} The quantitative results of predicted bounding boxes on the S3DIS and SUN RGB-D datasets are shown in Fig. \ref{fig:s3dis_show} and Fig. \ref{fig:sunrgbd_show}, respectively. In particular, from Fig. \ref{fig:sunrgbd_show}, we can see that our proposed SPGroup3D can even detect some unlabeled objects in SUN RGB-D.

\paragraph{Inference time.} The inference time are 93ms (VoteNet), 169ms (H3DNet), 178ms (GroupFree),101ms (FCAF3D), 268ms (CAGroup3D*) and 131ms (ours) on the same workstation (single NVIDIA RTX 3090 GPU card). 
Although our model is slightly slower than VoteNet and FCAF3D, our results on mAP@0.25 and mAP@0.5 of ScanNet V2 are 15.7 / 26.1 and 2.8 / 2.3 higher than theirs.

\paragraph{Per-class Evaluation.} In this part, we report the results for each category on ScanNet V2, S3DIS and SUN RGB-D. Tab. \ref{tab:25scan},  Tab. \ref{tab:50scan} report the results of 18 classes of ScanNet V2 at 0.25 and 0.5 IoU thresholds, respectively. Tab. \ref{tab:s3dis25},  Tab. \ref{tab:s3dis50} show the results of 5 classes of S3DIS on 0.25 and 0.5 IoU thresholds, respectively. Tab. \ref{tab:25sun},  Tab. \ref{tab:50sun} show 10 classes of SUN RGB-D results on 0.25 and 0.5 IoU thresholds, respectively. 

For Scannet V2 and S3DIS, comprehensive Tab. \ref{tab:25scan}, Tab. \ref{tab:50scan}, Tab. \ref{tab:s3dis25}, and Tab. \ref{tab:s3dis50},  we can see that the improvement of SPGroup3D is mainly reflected in small class, such as picture, and thin classes, such as door, curtain, window and board. This improvement is achieved by using the components we proposed in this paper.
For SUN RGB-D, although our model did not achieve the best results on mAP@0.5, we still achieve good performance in some classes, such as bed, bookshelf, desk and sofa.

\begin{table*}[h]
	\large
	\centering
	\renewcommand{\arraystretch}{1.5}{
		\resizebox{1.0\textwidth}{!}{
			\begin{tabular}{c|c|c|c|c|c|c|c|c|c|c|c|c|c|c|c|c|c|c|c}
				\toprule
				Methods & cab & bed & chair & sofa & tabl & door & wind & bkshf & pic & cntr & desk & curt & frig & showr & toil & sink & bath & ofurn & mAP \\
				\midrule
				\midrule
				VoteNet \cite{votenet} & 47.8 & 90.7 &90.0 &90.7 &60.2 &53.8 &43.7 &55.5 &12.3 &66.8 &66.0 &52.3 &52.0 &63.9 &97.4 &52.3 &92.5 &43.3 &62.9 \\
				MLCVNet \cite{mlcvnet} & 42.5 & 88.5 & 90.0 & 87.4 & 63.5 & 56.9 & 47.0 & 57.0 & 12.0 & 63.9 & 76.1 & 56.7 & 60.9 & 65.9 & 98.3 & 59.2 & 87.2 & 47.9 & 64.5\\ 
				BRNet \cite{brnet} & 49.9 &88.3 &91.9 &86.9 &69.3 &59.2 &45.9 &52.1 &15.3 &72.0 &76.8 &57.1 &60.4 &73.6 &93.8 &58.8 &92.2 &47.1 &66.1 \\
				H3DNet \cite{h3dnet} & 49.4 & 88.6 & 91.8 & 90.2 & 64.9 & 61.0 & 51.9 & 54.9 & 18.6 & 62.0 & 75.9 & 57.3 & 57.2 & 75.3 & 97.9 & 67.4 & 92.5 & 53.6 & 67.2\\
				GroupFree \cite{groupfree} &52.1 &92.9 &93.6 &88.0 &70.7 &60.7 &53.7 &62.4 &16.1 &58.5 &\textbf{80.9} &67.9 &47.0 &76.3 &99.6 &72.0 &\textbf{95.3} &56.4 &69.1 \\
				FCAF3D (Rukhovich et al. 2022) & 57.2 &87.0 & \textbf{95.0} & 92.3 &70.3 &61.1 &60.2 &64.5 &29.9 &64.3 &71.5 &60.1 &52.4 &\textbf{83.9} &\textbf{99.9} &\textbf{84.7} &86.6 &\textbf{65.4} &71.5 \\
				\midrule
				SPGroup3D (ours) & \textbf{58.0} &\textbf{88.2} &94.2 &\textbf{93.0} & \textbf{73.4} &\textbf{68.4} &\textbf{65.9} &\textbf{66.9} &\textbf{39.3} &\textbf{72.5} &79.6 &\textbf{64.2} &\textbf{64.0} &79.6 &99.8 &77.3 &90.2 &62.2 &\textbf{74.3}\\
				\bottomrule
	\end{tabular}}}
	\caption{3D detection scores per category on the ScanNetV2, evaluated with mAP@0.25 IoU.}
	\label{tab:25scan}
\end{table*}

\begin{table*}[h]
	\centering
	\renewcommand{\arraystretch}{1.5}{
		\resizebox{1.0\textwidth}{!}{
			\begin{tabular}{c|c|c|c|c|c|c|c|c|c|c|c|c|c|c|c|c|c|c|c}
				\toprule
				Methods & cab & bed & chair & sofa & tabl & door & wind & bkshf & pic & cntr & desk & curt & frig & showr & toil & sink & bath & ofurn & mAP \\
				\midrule
				\midrule
				VoteNet \cite{votenet} &8.1 &76.1 &67.2 &68.8 &42.4 &15.3 &6.4 &28.0 &1.3 &9.5 &37.5 &11.6 &27.8 &10.0 &86.5 &16.8 &78.9 &11.7 &33.5 \\
				BRNet \cite{brnet} & 28.7 &80.6 &81.9 &80.6 &60.8 &35.5 &22.2 &48.0 &7.5 &\textbf{43.7} &54.8 &39.1 &51.8 &35.9 &88.9 &38.7 &84.4 &33.0 &50.9 \\
				H3DNet \cite{h3dnet} & 20.5 & 79.7 & 80.1 & 79.6 & 56.2 & 29.0 & 21.3 & 45.5 & 4.2 & 33.5 & 50.6 & 37.3 & 41.4 & 37.0 & 89.1 & 35.1 & \textbf{90.2} & 35.4 & 48.1\\
				GroupFree \cite{groupfree} &26.0 &81.3 &82.9 &70.7 &62.2 &41.7 &26.5 &55.8 &7.8 &34.7 &\textbf{67.2} &43.9 &44.3 &44.1 &92.8 &37.4 &89.7 &40.6 &52.8 \\
				FCAF3D (Rukhovich et al. 2022) &\textbf{35.8} &81.5 &\textbf{89.8} &\textbf{85.0} &62.0 &44.1 &30.7 &58.4 &17.9 &31.3 &53.4 &44.2 &\textbf{46.8} &\textbf{64.2} &91.6 &\textbf{52.6} &84.5 &57.1 &57.3 \\
				\midrule
				SPGroup3D (ours) & 35.2 &81.7 &88.3 &84.5 &\textbf{64.8} &\textbf{51.1} &\textbf{40.4} &\textbf{63.9} &\textbf{26.9} &32.1 &60.1 &\textbf{47.9} &43.9 &63.3 &\textbf{97.5} &51.2 &84.3 &\textbf{53.6} &\textbf{59.6} \\
				\bottomrule
	\end{tabular}}}
	\caption{3D detection scores per category on the ScanNetV2, evaluated with mAP@0.50 IoU.}
	\label{tab:50scan}
\end{table*}

\begin{table*}[h]
	\centering 
	\renewcommand{\arraystretch}{1.5}{
		\resizebox{0.55\textwidth}{!}{
			\begin{tabular}{c|c|c|c|c|c|c}
				\toprule
				Methods & table & chair & sofa & bkcase & board & mAP \\ 
				\midrule
				\midrule
				GSDN (Gwak et al. 2020) & \textbf{73.7} & \textbf{98.2} & 20.8 & 33.4 & 12.9 & 47.8 \\
				FCAF3D (Rukhovich et al. 2022) & 69.7 & 97.4 & \textbf{92.4} & 36.7 & 37.3 & 66.7 \\ 
				\midrule
				SPGroup3D (ours) & 67.2 & 96.7 & 83.4 & \textbf{38.2} & \textbf{60.0} & \textbf{69.1} \\ 
				\bottomrule
	\end{tabular}}}
	\caption{3D detection scores per category on the S3DIS, evaluated with mAP@0.25 IoU.}
	\label{tab:s3dis25}
\end{table*}

\begin{table*}[h]
	\centering 
	\renewcommand{\arraystretch}{1.5}{
		\resizebox{0.55\textwidth}{!}{
			\begin{tabular}{c|c|c|c|c|c|c}
				\toprule
				Methods & table & chair & sofa & bkcase & board & mAP \\ 
				\midrule
				\midrule
				GSDN (Gwak et al. 2020) & 36.6 & 75.3 & \phantom{0}6.1 & \phantom{0}6.5 & 1.2 & 25.1 \\
				FCAF3D (Rukhovich et al. 2022) & 45.4 & 88.3 & \textbf{70.1} & \textbf{19.5} & 5.6 & 45.9 \\ 
				\midrule
				SPGroup3D (ours) & \textbf{47.2} & \textbf{89.5} & 61.3 & 14.6 & \textbf{23.9} & \textbf{47.2} \\ 
				\bottomrule
	\end{tabular}}}
	\caption{3D detection scores per category on the S3DIS, evaluated with mAP@0.25 IoU.}
	\label{tab:s3dis50}
\end{table*}

\begin{table*}[h]
	\large
	\centering
	\renewcommand{\arraystretch}{1.5}{
		\resizebox{0.8\textwidth}{!}{
			\begin{tabular}{c|c|c|c|c|c|c|c|c|c|c|c}
				\toprule
				Methods &bathtub &bed &bkshf &chair &desk &dresser &nstand &sofa &table &toilet &mAP \\
				\midrule
				\midrule
				VoteNet \cite{votenet} &75.5 &85.6 &31.9 &77.4 &24.8 &27.9 &58.6 &67.4 &51.1 &90.5 &59.1 \\
				MLCVNet \cite{mlcvnet} & 79.2 & 85.8 & 31.9 & 75.8 & 26.5 & 31.3 & 61.5 & 66.3 & 50.4 & 89.1 & 59.8\\ 
				H3DNet \cite{h3dnet} & 73.8 & 85.6 & 31.0 & 76.7 & 29.6 & 33.4 & 65.50 & 66.5 & 50.8 & 88.2 & 60.1\\
				BRNet \cite{brnet} &76.2 &86.9 &29.7 &77.4 &29.6 &35.9 &65.9 &66.4 &51.8 &91.3 &61.1 \\
				HGNet \cite{hgnet} & 78.0 & 84.5 & 35.7 & 75.2 & 34.3 & 37.60 & 61.7 & 65.7 & 51.6 & 91.1 & 61.6\\
				GroupFree \cite{groupfree} &\textbf{80.0} &87.8 &32.5 &79.4 &32.6 &36.0 &66.7 &70.00 &53.80 &91.10 &63.0 \\
				FCAF3D (Rukhovich et al. 2022) &79.0 &88.3 &33.0 &81.1 &34.0 &\textbf{40.1} &\textbf{71.9} &69.7 &53.0 &\textbf{91.3} &64.2 \\
				\midrule
				SPGroup3D (ours) &78.6 &\textbf{89.6} &\textbf{38.1} &\textbf{81.1} &\textbf{38.6} &37.8 &66.3 &\textbf{75.0} &\textbf{55.9} &90.3 &\textbf{65.1} \\
				\bottomrule
	\end{tabular}}}
	\caption{3D detection scores per category on the SUN RGB-D, evaluated with mAP@0.25 IoU.}
	\label{tab:25sun}
\end{table*}

\begin{table*}[t]
	\large
	\centering
	\renewcommand{\arraystretch}{1.5}{
		\resizebox{0.8\textwidth}{!}{
			\begin{tabular}{c|c|c|c|c|c|c|c|c|c|c|c}
				\toprule
				Methods &bathtub &bed &bkshf &chair &desk &dresser &nstand &sofa &table &toilet &mAP \\
				\midrule
				\midrule
				VoteNet \cite{votenet} &45.4 &53.4 &6.8 &56.5 &5.9 &12.0 &38.6 &49.1 &21.3 &68.5 &35.8 \\
				H3DNet \cite{h3dnet} & 47.6 & 52.9 & 8.6 & 60.1 & 8.4 & 20.6 & 45.6 & 50.4 & 27.1 & 69.1 & 39.0\\
				BRNet \cite{brnet} & 55.5 & 63.8 & 9.3 & 61.6 & 10.0 & 27.3 & 53.20 & 56.7 & 28.6 & 70.9 & 43.7\\
				GroupFree \cite{groupfree} &64.0 &67.1 &12.4 &62.6 &14.5 &21.9 &49.8 &58.2 &29.2 &72.2 &45.2\\
				FCAF3D (Rukhovich et al. 2022) &\textbf{66.2} &69.8&11.6 &\textbf{68.8} &14.8 &\textbf{30.1} &\textbf{59.8} &58.2 &\textbf{35.5} &\textbf{74.5} &\textbf{48.9}\\
				\midrule
				SPGroup3D (ours) &55.0 & \textbf{70.8} &\textbf{15.7} &66.3 &\textbf{15.1} &28.2 &53.2 &\textbf{61.7} &33.8 & 72.4 & 47.2 \\
				\bottomrule
	\end{tabular}}}
	\caption{3D detection scores per category on the SUN RGB-D, evaluated with mAP@0.50 IoU.}
	\label{tab:50sun}
\end{table*}

\end{document}